# A Hybrid Three Layer Architecture for Fire Agent Management in Rescue Simulation Environment

**Alborz Geramifard, Peyman Nayeri, Reza Zamani-Nasab & Jafar Habibi**
Department of Computer Engineering,
Sharif University of Technology, Tehran, Iran
(geramifa, nayyeri, zamani)@ce.sharif.edu, habibi@sharif.edu

***Abstract:** This paper presents a new architecture called FAIS for imple- menting intelligent agents cooperating in a special Multi Agent environ- ment, namely the RoboCup Rescue Simulation System. This is a layered architecture which is customized for solving fire extinguishing problem. Structural decision making algorithms are combined with heuristic ones in this model, so it's a hybrid architecture.*

## 1. Introduction

The RoboCup Rescue Simulation system makes a test-bed for implementation of various Multi-Agent algorithms. Its capabilities cover a wide range of possible styles of algorithms. It is also a standard environment for testing different techniques of making standard software agents with distributed architecture[10]. Rescue Simulation System also provides a standard framework for testing pro- posed algorithms and mathematical models of disaster events[8].
Designing an autonomous agent set like the one that is required for RoboCup Rescue Simulation is a little bit more of a challenge. Planning effective collab- oration for a Multi-Agent team in disastrous environments still remains a challenging area in AI. Efforts of Multi-Agent researchers have provided somewhat of a standard in modeling and designing software. A lot of effort has gone into reaching coordination between different agents and making autonomous deci- sions that work toward the team goal[9]. But practical results in complicated domains such as RoboCup Rescue Simulation indicate that heuristic criteria still remain as a major part of a successful system[11]. This may signal lack of satisfactory models for these complicated situations. These evidences encouraged us towards the implementation of a hybrid system called FAIS[1]. Our structured model constitutes the core of the system which acts on advices generated by heuristic components. In fact heuristic components decrease the complexity of the domain and the structured core analyzes these reasonable incoming advices and makes decision. Practical results convinced us that this is an achievement over pure heuristic designs in this domain. The next section will introduce a brief problem deffnition. Then the architecture will be presented (Section 3) and examined for Fire Brigade (Section 4) and Fire Station(Section 5) agents. In Section 6 we will provide some experimental results and finally in section 7 we will conclude the paper.

## 2. Problem Definition

In this section we consider a simplified version of fire extinguishing problem in RoboCup Rescue Simulation[7] . The system consists of several communicat- ing modules: There is a standard fire simulator that simulates the growth and damage of fire in the simulated city. The simulated city itself is represented and managed by a software process named GIS. Two types of agents can help extinguish fire: Fire Brigades and Fire Stations. In our version of the problem we consider only one instance of working Fire Station. There is a coordinating process that all other processes interact through it, so it is reasonable to call this process Kernel. Communication is accomplished via an extended UDP protocol named Long-UDP.

---

[1] Fire Agent Integrated System

Fire Brigades can obtain visual information of their vicinity through Kernel, also they can send move and extinguish commands. A Fire Station agent can get visual information but can't send move or extinguish command. There is a limited bandwidth for communication through kernel module. Fire Station is more capable of sending and receiving messages, so -as its name indicates- it can be considered as an ofiine decision making agent which can generate guidelines and advices for Fire Brigades. Fire Brigades also sufier damage during the simulation process while they are in fiery buildings. Every time a Fire Brigade extinguishes a fire, its available water decreases. This simulates existence of water tank for the agent. Moreover, there are certain buildings in the city that agents can heal themselves in them (decreasing the amount of damage) or fill their water tanks (increasing the amount of available water) while they are inside them. These buildings are called Refuges. Detailed information is available in [7].

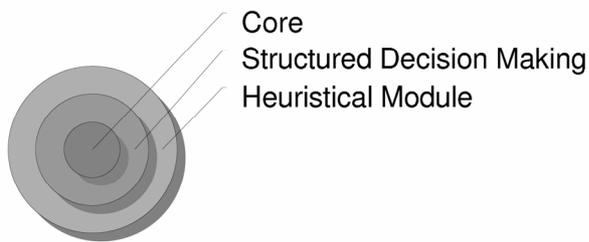

Fig. 1. Three layered architecture

The goal of the simulation is to build a system of agents that can extinguish the simulated city and decrease the ratio of the burnt buildings to total buildings as much as possible in the simulation deadline.

### 3. FAIS Architecture

Figure 1 shows a schematic view of the agents' architecture. This structure is used in all agents. According to each agent's task, these layers will be customized. The system consists of three main layers. The lower level is responsible for analyzing sensory data and generating several alternatives. In fact the output of this layer makes the customized agent model of the world. This layer is named Heuristic Module in figure 1. Decision Making layer is placed upon extracted alternatives of the previous layer. The minimized world model provided by the heuristic modules is the in- put of more structured algorithms. There are a wide range of algorithms that cover the topic of decision making. Most of these algorithms tend to make deci- sions upon a restricted number of well-defined parameters. The domain of the rescue simulation contains a large number of important parameters that should participate in decision making. Experiments for making direct decisions through passing all these parameters to such algorithms have shown unsatisfactory re- sults. So, the role of heuristic components in this design is to provide an abstract view of the world model as an input for decision making routines. Finally there is a Core layer on top of the architecture. All decisions made by the underlying algorithms should be passed through this layer. This layer acts like a filter and applies some predefined rules from a certain database to the input decisions. This is because structured algorithms are generally acceptable, but most times there are a set of situations that these algorithms fail to behave well. The core layer is a standard place for handling these special cases. These cases include situations such as when there are no visible fiery buildings or when an agent is trapped in a section of the city. In such situations the Core Layer makes the final logical decision.

### 4. Fire Brigade Design

We are going to introduce the details of the this three-layered architecture in the Fire Brigade agent.
Fire Brigade's lower layer consists of the following modules:

- BFS: This is the system's path planning module. Typically the agent needs to find physical paths on the simulated city to change its position. The input to this problem is a graph. Graph edges represent the open roads at the beginning of the disaster simulation. As the time goes by, some roads will be opened so this graph is a dynamic graph. The problem of handling such a dynamic graph is solved partially by deterministic approaches like continuous Dijkstra[1]; however since positions of blocked roads are unknown till late in the simulation, such a method will not give a satisfactory result. The base of our solution here is a simple BFS over a graph of all the potentially open roads. As the agents explore the graph, closed roads will be discovered and marked on the graph.

- Collision Detection: Sometimes agents can't travel all of the roads of an open path. This occurs for example when a certain direction of the road is not accessible because some other agents have blocked the way. In this situation this heuristic strategy is used: Upon discovery of a blocked direction of a road, the certain edge will be removed from the graph, temporarily. After a fixed period of time, the edge will be tagged as an open road again.

- Feedback: At certain intervals, each agent reports some important information that have been discovered during its explorations to the Fire Station agent . This complementary information is the base of what is known as station's world model.

- Critical Injury: When an agent receives a certain minimum level of damage, it ignores all plans except going to Refuge buildings.

- Water InsuÆciency: When a Fire Brigade agent detects that it has ran out of water, all plans will be ignored except going to the nearest Refuge building. Figure 2 shows the total view of these components in respect to the total ar- chitecture. The following components are deterministic decision making modules in the Fire Brigade structure.

Ordered Based Behavior: This is a component that makes the main agent decisions. This sub-layer works according to an advice based structure[2]. The Fire Station processes the world model, which is mostly provided by the Fire Brigades' feedbacks, and generates the advices that Fire Brigades obey in order to remain coordinated.

Wander Mode: When Fire Brigades extinguish all of the fires, they start to explore the city regularly. This seems to be an effective method, since extra information about the city will be discovered through the remaining time of the simulation.

$$V(b) = \beta N(b) + \gamma HV(b) - \alpha \sum_{i \in J} d_i(b) \qquad (1)$$

$d_i(x)$ is the distance between Fire Brigade agent number i and building x. J is the set of indices of currently free agents.

$N(x)$ is the number of unfired neighbors of the building x. $HV(x)$ is a value associated with a building x that indicates how much x can be destructive if we let it spread the fire through the city. For example, a fiery building in the margin of the city is much less dangerous than a fiery building in the city center. Figure 4 shows the details. First all fiery neighbors are computed, and the convex hull of the set, H, is constructed through a Graham scan method [3]. Then a unit vector u is computed such that in position of building x, it points toward bisector of angle of H at vertex x (see figure 4).

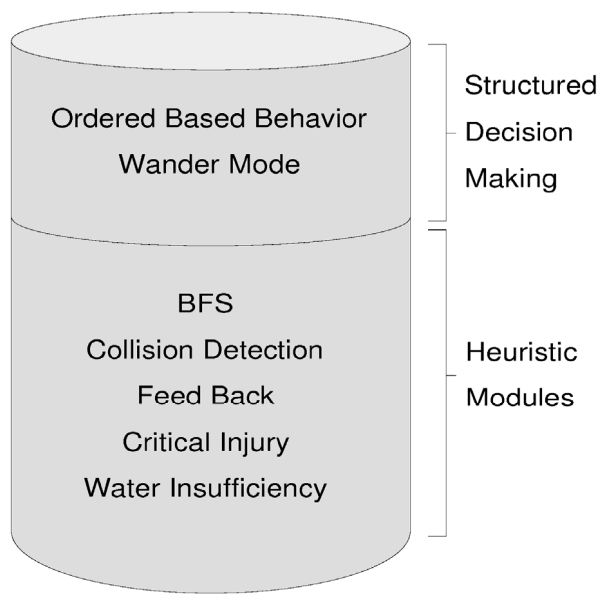

Fig. 2. Fire Brigade structure

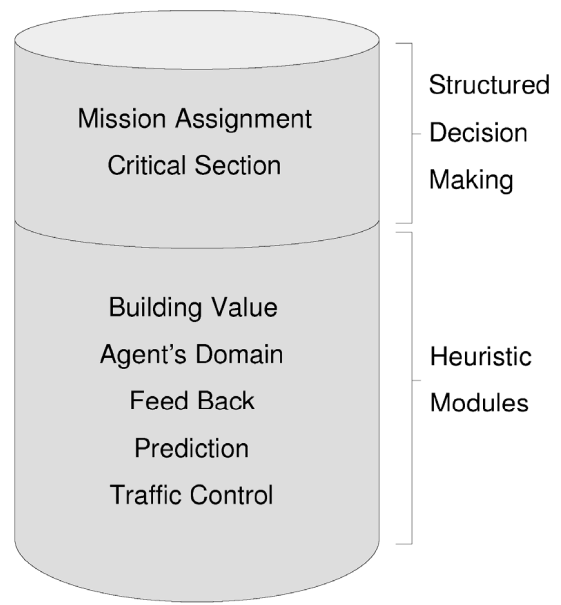

Fig. 3. Fire Station Design

## 5. Fire Station Design

Fire Station module is responsible for centered decision making and coordination. It computes a profitable subset of fiery buildings and assigns a subset of idle agents to them for the extinguishing operation. The system sets a proper timeout for assigned mission called mission time, and after this deadline, Fire Station considers the assigned Fire Brigades as free agents. Fire Brigades will also abandon the mission if the timeout expires. Figure 3 shows the details of the Fire Station. The lower layer components are: Building Value: This is a heuristic real value, computed for each building. Buildings with bigger values are candidates for extinguishing in the mission assignment. The value for building b is computed in the following way:

All other fiery buildings that are not much further than a certain margin, from the semi-line in direction of u starting at x, are examined to find the nearest one to x, let y be that building. We define $HV(x) = distance(x; y)$. $\alpha$, $\beta$ and are positive factors that are determined after fine tuning of implemented agents. Since a small difference in the score result could have been caused by the random functions involved during the simulation process we couldn't apply the fine tuning process by automatic means. Furthermore the final score doesn't show if each agent was acting in a more logical way during different parts of the simulation, so we changed the parameters by viewing the logs and deciding whether the agents had acted better or not. Because we changed these parameters by visually surveying the logical behavior of our agents they are independent of the city. The intuition behind (1) is that buildings with more unfired neighbors, hav- ing more dangerous position to spread fire and with much free Fire Brigades near them are more profitable to choose.

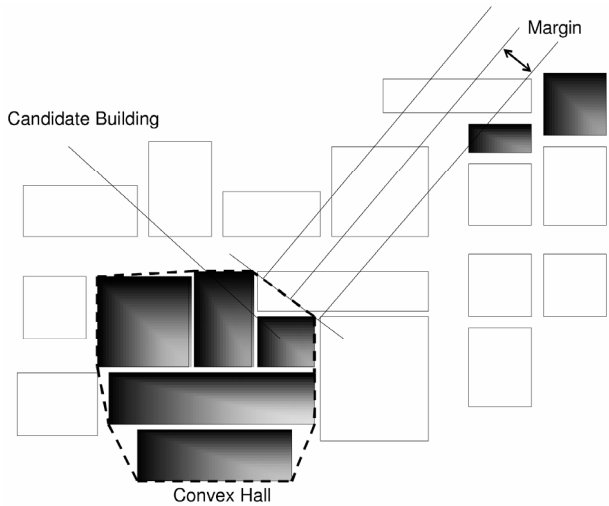

Fig. 4. Finding the fire border

Agent's Domain: For the previous section, we should compute the set of free agents for each building. But this differs for each building, since the roads graph is dynamic and some idle agents can't reach some certain buildings (that is some certain graph vertices). So, in this component, the set of reachable buildings for each agent is computed through a BFS circumnavigation. Therefore, the set of free agents considered for each building can be a proper subset of all idle agents.

Prediction: The Fire Station needs to have proper approximation of future situation to schedule effective missions. Since phenomena like fire spreading in deterministic environments obey some certain rules, it is possible to provide some approximation. This module is responsible for maintaining a compact set of previously known scenarios and matching the current scenario with them to predict some future parameters like which buildings will be ignited in the next cycles and how much water is sufficient for extinguishing a certain building. A feed forward neural network provides the compact representation in addition to proper response time and accuracy [4].

Feedback: The process of fire spreading is dependent on the current status of the city. For example broken buildings are more vulnerable to fire and therefore spread the fire more quickly. The predictor must be informed of such information in order to predict properly. In fact Feedback module of Fire Brigade provides discovered information as agents explore the city, and this module prepares the received information for the prediction module.

Traffic Control: When agents are assigned to missions, in many cases the roads are blocked because of working agents. Since agents are working independently, it is very difficult for them to resolve this problem on their own. This module of Fire Station monitors the assigned missions and makes proper advices for each agent in order to avoid road blocks as much as possible. In many cases, sending agents to the buildings near the target building is more appropriate than keeping them in the roads. Figure 5 shows a sample of this event before using Traffic Control module (left) and after applying Traffic Control rules (right).

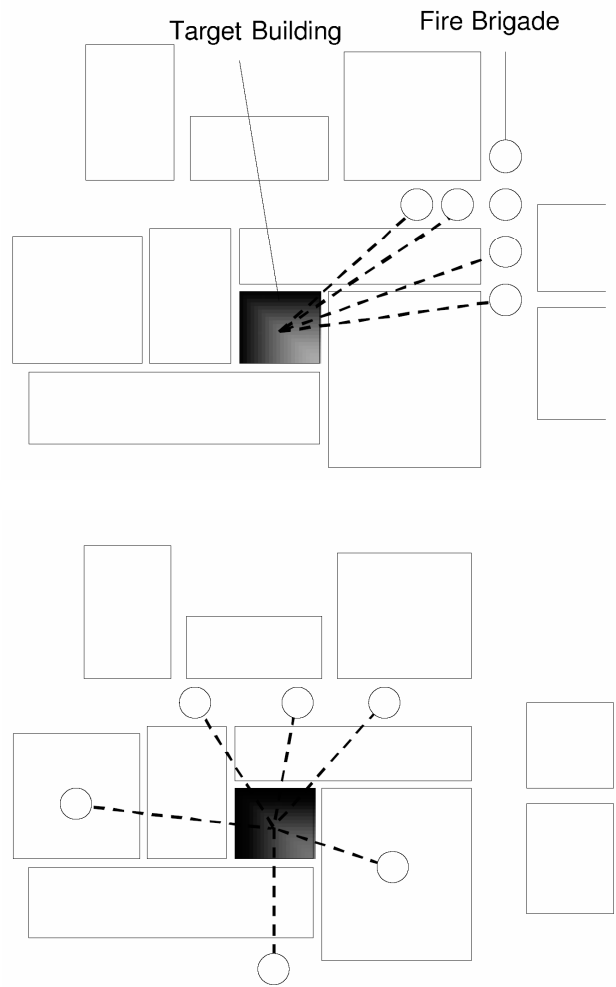

Fig. 5. Fire Station Traffic Control

The structured decision making components are:

Mission Assignment: This module does the real task of mission assignment. At first prediction module approximates future state of the scene and then potential missions are identified. Building Value module then selects some profitable buildings between these. This comes up with a limited number of proper buildings. Free agents should be assigned to these buildings in an optimum manner, i.e. sum of the distances for all agents to reach their mis- sions be minimum. This problem is a restricted type of the graph matching problem with limited degrees at vertices [6]. An LP solver does these final assignments [5].

- Critical Section: The way our agents behave in the first cycles of the simula- tion is of utmost importance. Extinguishing more buildings in the first cycles will reduce the amount of fire spreading quite significantly. This will make handling the disaster far more easier. Furthermore, not only the prediction module hasn't enough information to make satisfactory predictions in the first cycles, but also if the prediction results are acceptable, many roads are blocked at the start of simulation, so many assigned missions will not be completed successfully. Critical Section component overrides mission assign- ment in the certain interval of initial cycles. In this interval, missions are not assigned optimally, but simply all agent are assigned to the best building for extinguishing. This has shown a better performance than using prediction and assignment from the beginning.

**6. Experimental Results**

A set of Fire Agents (Fire Brigade, Fire Station) based on FAIS architecture is fully implemented and participated as Arian team in the 2003 Rescue Simulation Robocup Competitions. There was a wide range of strategies available in the competition and our agents showed the best performance between them and placed first. Table 1 is the result of the first round of the competition. In this section we will discuss how FAIS architecture helped us in achieving this result.

Note: Since all of the teams didn't release their actual codes, these results are based solely on visual comparison of the agent's behaviors during the competition.

*6.1 An overall Survey*
Critical Section Since in some sections of the competitions there were maps with lots of initial fiery buildings, correct decision on task assignment in the first few cycles of the simulation greatly affected the overall result.

Prediction In cities with lots of initial fire points teams had problems with their agents losing control of the fire sites. In such cases our predictor module proved to be quite useful since it provided our Fire Station with a good overall view of the situation. This helped our agents to work in a logical manner even during such circumstances.

Centralized Decision Making Most of the teams focused their algorithms on their Fire Agent so the low degree of cooperation between their agents caused them to work independently and extinguish buildings which weren't quite use- ful. The centralized decision making led us to have agents with eÆcient water consumption. In cities which the refuges where inaccessible during some cycles of the simulation process, this eÆciency played a great role in our good results.

Communication The logs of some teams led us to believe that their Fire Stations only worked as a message conveyor rather than a decision maker. This caused their agents to work separately and not concentrate their power on the vital areas of the city.

TraÆc Control As mentioned before traÆc control is of utmost importance during this simulation process. In the logs we saw many teams had trouble with traÆc jams which caused major problems for their agents. The path planning process which our agents used help them in choosing optimum paths with lower traÆc. This process took advantage of our fire station to inform the agents of any temporary road blocks which occurred during the process.

| ARIAN | 62.35 |
|---|---|
| S.O.S. | 56.08 |
| The Black Sheep | 43.83 |
| YowAI2003 | 41.30 |
| Eternity | 38.74 |
| POLITECS2003 | 31.70 |
| NITRescue03 | 29.56 |
| RESQ FREIBURG | 28.58 |
| SBCE SAVOUR | 28.18 |
| RAYAN | 27.74 |
| RoboAkut | 25.28 |
| PAKRescueTeam | 23.50 |
| SBCE RES | 22.71 |
| Ferdowsi | 14.88 |
| UVA RESCUE C2003 | 13.56 |
| ToosRes | 13.50 |
| BanzAI | 10.12 |

Table 1. RoboCup 2003, Rescue Simulation Competition results, Round one

*6.2 Why FAIS?*
As mentioned in the overall survey the heuristic functions which we used, greatly helped us in achieving a good result. At first we developed our agents by using these functions as their heuristic modules but when the number of these functions increased we saw that the separate usage of these functions without a high level module which combines their outputs wouldn't have a good result. So we implemented a new layer in our architecture called Structured Decision Making which receives the results of these heuristic modules and applies them in the best possible way. In some circumstances the workow of the the agents needs to be changed, Core Layer as the highest level of this architecture decides that what major role should each agent play. By using this architecture the overall performance of our agents increased by a great degree.

## 7. Conclusion

In this paper we discussed a new architecture, which has been used to develop fire agents for the rescue simulation environment. Multi-agent environments are typically complicated and require a great degree of cooperation between the agents and a potent architecture in order to accomplish their duties. The proposed architecture by benefiting from its three layered design, overpowered any previously known and implemented architecture in the rescue simulation field. Future works will include testing this architecture on the police and ambulance agents of the rescue simulation environment.

**Acknowledgement**
This research was conducted in the RoboCup Laboratory of Sharif Univesrity of Technology and we would like to thank all the members of the Arian team who worked on the rescue simulation project.